\pdfoutput=1

\documentclass[11pt]{article}

\usepackage[preprint]{coling}
\usepackage{amsmath}
\usepackage{amsfonts}

\usepackage{times}
\usepackage{latexsym}
\usepackage{amssymb}

\usepackage[T1]{fontenc}
\usepackage[utf8]{inputenc}

\usepackage{microtype}

\usepackage{inconsolata}

\usepackage{graphicx}

\title{Optimizing Chain-of-Thought Reasoning: Tackling Arranging Bottleneck via Plan Augmentation}

 \author{Yuli Qiu, Jiashu Yao, Heyan Huang, Yuhang Guo\\
         School of Computer Science and Technology, Beijing Institute of Technology, Beijing, China\\
         \texttt\{yuliqiu, yaojiashu, hhy63, guoyuhang\}@bit.edu.cn}

\begin{document}
\maketitle
\begin{abstract}
Multi-step reasoning ability of large language models is crucial in tasks such as math and tool utilization. Current researches predominantly focus on enhancing model performance in these multi-step reasoning tasks through fine-tuning with Chain-of-Thought (CoT) steps, yet these methods tend to be heuristic, without exploring nor resolving the bottleneck. In this study, we subdivide CoT reasoning into two parts: arranging and executing, and identify that the bottleneck of models mainly lies in arranging rather than executing. Based on this finding, we propose a plan-based training and reasoning method that guides models to generate arranging steps through abstract plans. We experiment on both math (GSM8k) and tool utilization (ToolBench) benchmarks. Results show that compared to fine-tuning directly with CoT data, our approach achieves a better performance on alleviating arranging bottleneck, particularly excelling in long-distance reasoning generalization.

\end{abstract}

\section{Introduction}
Multi-step reasoning ability is prevalent in large language model (LLM) applications, which involve math problems\cite{cobbe2021training,luo2023wizardmath}, tool utilization\cite{schick2024toolformer,li2023api} and so on. During multi-step reasoning, the model utilizes Chain-of-Thought (CoT) reasoning\cite{wei2022chain} to analyze and derive the final answer, crucial for decomposing complex tasks into manageable sub-reasoning steps.‌ Previous studies focus on enhancing the accuracy of CoT reasoning steps\cite{yao2024,besta2024graph,yuan2023scaling}. 
However, these approaches are mostly heuristic, and lack comprehensive interpretable analyses of the bottlenecks that affect CoT performance in multi-step reasoning.

\begin{figure*}[t]
	\includegraphics[width=1\linewidth]{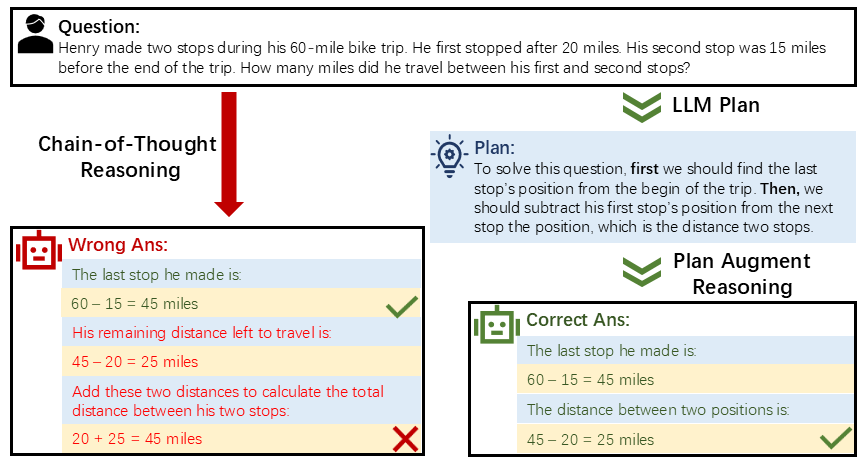} \hfill
	\caption {Examples of a math multi-step reasoning problem. This task can be split into two parts: \textbf{Arranging}(blue box) and \textbf{Executing}(yellow box). In the current \textbf{Chain-of-Thought-Reasoning}, these two parts are interwoven with each other, which may lead to errors in subsequent reasoning. We find the bottleneck of multi-step reasoning task is arranging rather than executing, and propose a method that allows LLM to first generate abstract arranging\textbf{(plan)} and use \textbf{Plan Augment Reasoning} to guide subsequent reasoning steps, ensuring the reliability of each step.}
	\label{example}
\end{figure*}

To address the lack of bottleneck analysis, we further divide CoT reasoning of these tasks into two parts: arranging and executing. In arranging, the model identifies the goal of each sub-step (what to do), while executing is responsible for specific completion (how to achieve). We focus on multi-step reasoning tasks such as math and tool utilization that involve instantiated calculations or invocations. As a math problem shown in Figure~\ref{example}, arranging gives the solution goal of next step, while arithmetic calculations are in executing.

From the perspective of the division of CoT, we propose a new method on evaluating and analyzing multi-step reasoning ability in math to find the bottleneck. Through theoretical derivations and empirical study, our analysis of selected models indicates that most recent models performs well in executing like simple arithmetic calculations, and arranging is the primary bottleneck. To further validate the arranging bottleneck, we employ arranging steps planned by GPT-4\cite{Achiam2023GPT4TR} to replace other model's arranging process, and we find that the multi-step reasoning ability has been significant improved.

To address the arranging bottleneck, we propose a plan-based method, which includes plan augment reasoning and plan-centric supervised fine-tuning (SFT). In our work, plan can be seen as an abstract collection of arranging steps (without executing) in CoT (Figure~\ref{example}), such as solving idea of math question, or task decomposition and API selection in tool utilization\cite{qiao2024autoact}. In plan augment reasoning, we utilize plan to guide model reason along the correct path. In plan-centric SFT, we set both plan and CoT steps generation as training objective to enhance the  arranging. 

We experiment our method on two multi-step reasoning tasks: math problem and tool utilization. Compared to previous works which fine-tuned on exclusively CoT steps, our plan-based method achieves significantly better results in these two tasks($p<0.01$), particularly excelling generalization in long-distance reasoning. These results show that our method can better enhance reasoning ability of models and alleviate the arranging bottleneck. The main contributions of this paper are as follows:
\begin{itemize}
\item We propose a new perspective on CoT reasoning steps, which is divided into two interleaved parts: arranging and executing.

\item We make analysis on model's multi-step reasoning ability in math problem based on our division, finding that arranging rather than executing is the bottleneck of models.

\item We propose a plan-based training and reasoning method, which can better alleviate the arranging bottleneck in multi-step reasoning.
\end{itemize}

\section{Related Works}
\subsection{Chain-of-Thought Reasoning}
Chain-of-Thought (CoT)\cite{wei2022chain} plays an important role in reasoning tasks with LLMs. \citet{NEURIPS2023_dfc310e8} show that a content size auto-regressive transformers is unable to solve basic arithmetic or equation tasks directly without CoT through theoretical analysis, which provides a perspective on the mechanism of CoT.

More researchers focus on enhancing the CoT reasoning of LLMs. Besides prompt strategy\cite{zhang2022automaticchainthoughtprompting}, some researchers alter the linear CoT reasoning to multi reasoning paths sampling, such as CoT-SC\cite{Wang2022SelfConsistencyIC}, ToT\cite{yao2024} and GoT\cite{besta2024graph}, to enable the model explore correct reasoning path step by step. Unlike the above studies, researchers also design multi-stage CoT frameworks for specific tasks, including log parsing\cite{zhang2024lemurlogparsingentropy}, code generation\cite{li2023structuredchainofthoughtpromptingcode}, multi-modal detection\cite{xu-etal-2024-exploring}, and so on.

LLMs can also improve reasoning ability by fine-tuning on CoT steps\cite{magister-etal-2023-teaching}, which can be acquired through methods such as rejection sampling\cite{yuan2023scaling}, model distillation\cite{liang-etal-2023-gpt} and self-instruct\cite{wang2022self}.

The current researches on CoT primarily focus on enhancing its accuracy through prompt engineering, designing reasoning structures or fine-tuning on CoT steps. However, there has been a lack of in-depth analysis regarding its underlying mechanisms and bottlenecks. Different from the above, we delve into the CoT process of multi-step reasoning, identify the arranging bottleneck and specifically propose a CoT training and reasoning method based on plan rather than detailed CoT steps.

\subsection{Planning with LLMs}
LLMs have shown impressive performance on multi-step reasoning tasks with the help of planning ability, which are exploring by many researchers. For example, \citet{momennejad2024evaluating} draw on the research in cognitive science and use cognitive map to test model's planning abilities, while \citet{valmeekam2023planning} assess model's performance in generating PDDL to guide specific task solving. Other researchers focus on more practical tasks, such as evaluating models' planning ability in tool utilization\cite{chen2024t}, fine-tuning a language model to generate tool calling plans\cite{qiao2024autoact} or code generation plans\cite{sun2024unicoder}. 

These studies have already noticed the importance of plan in multi-step reasoning. However, they predominantly treat plan as an isolated component for evaluation or training, and lack uniform definitional standards. Our research regards plan as an abstraction of model's arranging steps. We derive plan from CoT steps and employ a specifically trained model to generate plan that will guide subsequent reasoning process.
\section{Bottleneck Analysis of Multi-Step Reasoning Task}
In this section, we aim to make an in-depth analysis on bottleneck of models in multi-step reasoning. We focus on a typical case study: math problem. Math ability of models can be separated into math reasoning and arithmetic calculation\cite{yang2023gpt},  corresponding to arranging and executing respectively. On this basis, we intend to explore whether the current primary bottleneck of math problem is in reasoning or calculation.

\subsection{Theoretical Analysis and Evaluation Methods}
A naive idea of evaluating calculation and reasoning ability is to directly count the number of these two types of errors in generated CoT steps. However, arithmetic expressions generated by models are often not standardized, making it difficult to extract by regular expressions. To address this, \citet{li2024common} utilize GPT-4 to distinguish error type. Nevertheless, this method is costly and there could be some concerns regarding its reliability. Therefore, we propose another evaluation strategy that does not rely on additional models.

Given a multi-step reasoning task $q$, LLMs employ CoT to arrive at final answer. As shown in Figure~\ref{example}, these steps can be seen as a series of interwoven arranging and executing. We can abstract these sub-arranging steps together as plan, which can be used to guide task completion.

Suppose math question $q$ involves $n$ arithmetic steps, and each step's accuracy is $p_{exe}$. Hypothesize that the probability of deriving correct outcome $Acc_q$ is contingent upon two conditions: the correctness of overall reasoning path, and the accuracy of each calculation step, which is actually a cumulative power-law of $p_{exe}$:
\begin{equation}
	\label{eq6}
	Acc_q \leq (p_{exe})^n
\end{equation}

Suppose $N$ denotes the set of possible step values of math problems in the dataset. The expected final arithmetic accuracy can be expressed as:
\begin{equation}
	\label{eq7}
	ExeAcc = \mathbb{E}_{n \in N}\left[ (p_{exe})^n \right]
\end{equation}
$ExeAcc$ measures arithmetic ability of models, without considering reasoning. 

To evaluate reasoning ability, we define $ReasonScore$ as:

\begin{equation}
	\label{eq8}
	ReasonScore = \frac{Acc}{ExeAcc}
\end{equation}
It is evident that $ReasonScore=1$ only when generated reasoning steps are completely right. A higher $ReasonScore$ indicates that the primary bottleneck is calculation, whereas a lower value indicates the bottleneck lies in reasoning.

To further validate the bottleneck, we replace model's arranging steps of question with correct arranging sequences as plan, which can guide model reason along the right path. In our opinion, this approach could bridge the gap in math reasoning among LLMs.

\subsection{Experiment Settings}
\begin{figure}[htb]
	\includegraphics[width=0.9\columnwidth]{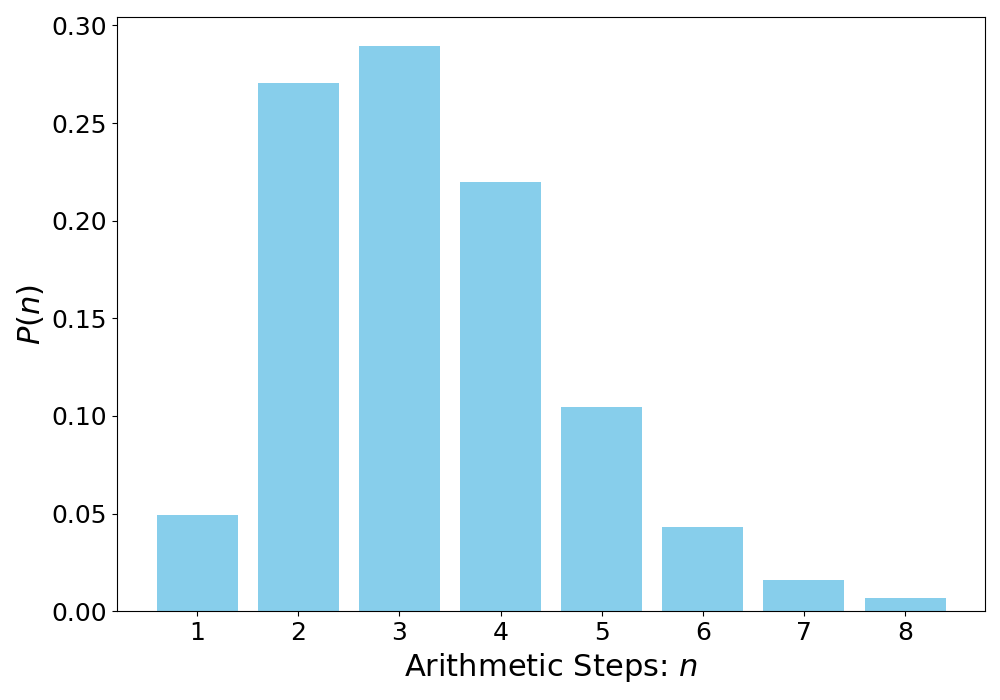}
	\caption{The distribution of arithmetic steps in GSM8k testset. Most of the problems need 2$\sim$4 steps to solve.}
	\label{gsm8k}
\end{figure}

\begin{figure}[htb]
	\includegraphics[width=0.9\columnwidth]{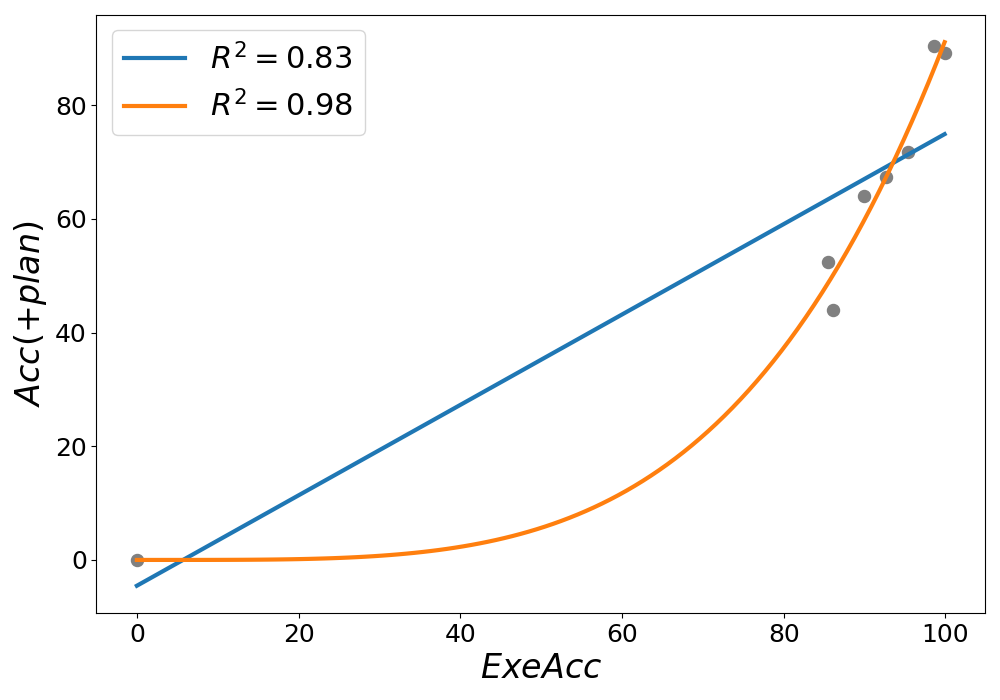}
	\caption{The results of the power-law fitting of experimental data $ExeAcc$ and $Acc(+plan)$ are closer than those of linear fitting, which indicates that model presents a power-law accumulation between math problems' final accuracy and single-step accuracy.}
	\label{fit}
\end{figure}

We conduct our experiments on GSM8k dataset\cite{cobbe2021training}, which consists of 7.4k training instances and 1.3k test instances. We extract 2,856 arithmetic formulas from the testset after deduplication. The distribution $P(n)$ of arithmetic steps for GSM8k is illustrated in Figure~\ref{gsm8k}. Considering that GPT-4-Turbo\footnote{https://platform.openai.com/docs/models/gpt-4-turbo-and-gpt-4} achieves an accuracy exceeding 90\% on GSM8k (Table~\ref{tab3}), we employ it to generate solution plan of questions, which ensure correctness of generated plans.

We select LLMs include GPT-3.5-turbo/GPT-4\cite{Achiam2023GPT4TR}, Qwen1.5-1.8b/4b/72b\cite{qwen1.5}, Llama2-7b/13b\cite{touvron2023llama}, and phi-2 (2.7b)\cite{javaheripi2023phi} in the experiment, covering main open-source and closed-source models of various parameter scales currently available. We test the reasoning and calculation abilities of above models in order to find their bottlenecks.

\subsection{Results}
\begin{table*}
	\centering
	\begin{tabular}{cccccc}
		\hline
		\textbf{Model} & \textbf{$p_{exe}$} & \textbf{$ExeAcc$} & \textbf{$Acc$} & \textbf{$ReasonScore(\%)$} & \textbf{$Acc(+plan)$} \\ \hline
		Qwen1.5-1.8b   & 86.1            & 59.5               & 36.9         & 62.0                & 44.0                \\
		phi2-2.7b      & 92.7            & 74.5               & 57.2         & 76.8                & 67.4                \\
		Qwen1.5-4b     & 95.4            & 81.4               & 57.1         & 70.2                & 71.8                \\
		Llama2-7b      & 85.5            & 58.3               & 26.8         & 46.0                & 52.4                \\
		Llama2-13b     & 89.9            & 67.8               & 38.8         & 57.2                & 64.0                \\
		Qwen1.5-72b    & 98.6            & 90.5               & 86.1         & 95.1                & 90.4                \\
		GPT-3.5-turbo  & 99.9            & 94.4               & 77.3         & 81.9                & 89.2                \\
		GPT-4          & 100             & 100                & 94.7         & 94.7                & -                   \\ \hline
		\textbf{Std.}            & 5.54            & 14.86              & 23.14        & 16.44                & 16.04               \\ \hline
	\end{tabular}
	\caption{Testing of the primary model's calculation and reasoning abilities on GSM8k. \textbf{Acc(+plan)} denotes the model's performance when aided by GPT-4 planning, and \textbf{Std.} represents the standard deviation among models. We only use GPT-4-generated plan to guide the reasoning of weaker models.}
	\label{tab3}
\end{table*}

Table~\ref{tab3} displays result of our experiment. It is clear that $Acc$ of all models is much lower than that of $ExeAcc$, which indicates that model make many reasoning mistakes. Besides, the $ReasonScore$ among models vary greatly, and most of them have a large gap with 100\%. All these show that models perform well in calculation, and reasoning is the bottleneck for weak models. Moreover, by substituting the reasoning process of GPT-4 for that of other models, the accuracy $Acc(+plan)$ has been significantly enhanced ($p=0.004$), which approximate $ExeAcc$. This indicates that the arranging part of CoT is separable. Although the results of two are approximate, the gap between them is might due to the difference in instruction following of models.

We also find that larger models tend to exhibit stronger reasoning abilities. As within the same model series (Qwen1.5, Llama2, and GPT families), larger models achieve higher $ReasonScore$. Larger models are often accompanied by more training data to fit\cite{Kaplan2020ScalingLF}, which improves the reasoning ability, while increase in calculation is relatively small.

Finally, as shown in Figure~\ref{fit}, we fit the relation between $Acc(+plan)$ and $ExeAcc$. The results indicate a power-law relation ($y=cx^a$) rather linear relation, confirming our hypothesis in Equation~\ref{eq6}.
\section{Plan-based Training and Reasoning Method}
We have taken math as an example in previous analysis experiment, and found that arranging rather than executing is the bottleneck of model in multi-step reasoning. To address this issue, we propose a plan-based training and reasoning framework to specifically enhance the arranging ability, thereby improving the performance of models in multi-step reasoning tasks.

\begin{figure}[h]
	\includegraphics[width=\columnwidth]{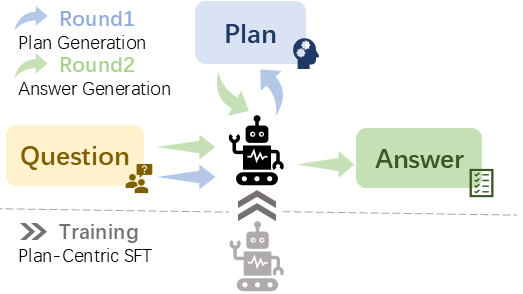}
	\caption{A framework of our method. Above the dotted line is Plan Augment Reasoning. LLM first generates plan in Round1, then make multi-step reasoning based on the plan to get final answer. Below the line is our Plan-Centric SFT to augment plan generation.}
	\label{role}
\end{figure}
\subsection{Plan Augment Reasoning}
For multi-step reasoning question $q$, previous works utilize model $M$ to generate CoT steps to arrive at the final answer $a$, which can be denoted as $P_M(a|q)$. Motivated from our experiment results in Table~\ref{tab3}, which conclude that using plans generated by GPT-4 to guide reasoning can significantly improve performance, we are motivated to enhance reasoning through plans.

As shown in Figure~\ref{role}, initially we prompt model to give the plan $p$ that excludes specific implementation details of question (Round1), which can be denoted as $P_M(p|q)$. In the context of math problems, the generated plan encompasses the general approach to solving the problem, devoid of explicit arithmetic expressions; in the realm of tool utilization, the model is required with decomposing the task and suggesting potential APIs on each step, without addressing parameter passing. 

In the stage of answer generation (Round2), we prompt model with both original question $q$ and generated plan $p$ together, enabling the model to generate intermediate arranging steps outlined in the plan, and pay more attention to executing such as arithmetic calculations and parameter passing. This process can be denoted as:
\begin{equation}
	\label{eq1}
	P_M(a|q)=P_M(p|q)P_M(a|p,q)
\end{equation}
Question $q$ and answer $a$ are originated from dataset; plan $p$ is generated by language model and serves as collection of sub-arranging steps in CoT, consequently directing the correct reasoning path.

\subsection{Plan-Centric SFT}
High-performance LLM (such as GPT-4) can directly generate  correct plans to guide subsequent reasoning (Table~\ref{tab3}). However, weaker models may generate incorrect arranging sequences as plan, which misdirect reasoning steps. Therefore, we intend to train models to generate correct plans. Different from previous training methods that utilize mixed arranging and executing CoT steps, we exclude specific execution steps and focus on abstract plan generation, which may help generalization. The objective of our plan-centric SFT includes two aspects: 
\begin{itemize}
	\item Given a specific question, generate an abstract plan that guides following reasoning steps.
	\item Given a specific question, generate arranging-executing CoT steps to get final answer.
\end{itemize}
The plan-centric SFT objective $L_{all}$ can be described as:
\begin{equation}
	\label{eq2}
	L_{all}=L_{plan}+L_{ans}
\end{equation}

Define $N$ is the total number of samples in the train set. The objective of generating plan $L_{plan}$ 
is represented as:
\begin{equation}
	\label{eq3}
	L_{plan}=-\sum_{n=1}^{N}log P(p|q)
\end{equation}
notations $p$ and $q$ represent plan and question respectively.

Similarly, the objective of generating final answer $L_{ans}$ 
is represented as:
\begin{equation}
	\label{eq4}
	L_{ans}=-\sum_{n=1}^{N}log P(a|q)
\end{equation}
notations $a$ and $q$ represent answer and question respectively. Actually, $L_{ans}$ presents the current training on CoT, focusing on both arranging and executing process. This approach also serves as our plan-based method's baseline. 

The plan-centric SFT is based on our hypothesis that incorporating $L_{ans}$ as part of our training will potentially enhance plan generation of the model when built upon foundational training provided by $L_{plan}$. To validate it, we experiment with setting the training objective to generate plans only, i.e., $L_{all}=L_{plan}$. The results are presented in our ablation study.

\section{Experiments}
\subsection{Dataset}
We use GSM8k benchmark for the training and evaluation of multi-step reasoning in math. Similar to our analysis experiment, we adopt synthetic data generation to acquire abstract plans for these problems: prompting GPT-4-Turbo to generate plans without specific calculations given problem $q$.

In tool utilization, we use ToolBench\cite{guo2024stabletoolbench} for training and evaluation. The data within ToolBench includes tasks, a list of executable APIs, multi-turn calling processes with feedback, and final responses. In this context, we conduct data cleaning and filtering on the multi-turn calling processes, retaining only successful API calls (including API names and parameters). Tasks in our dataset consist of two or more distinct API calls, which constitutes the multi-step reasoning we require. Besides, the authors in ToolBench accomplish tool selection by training an independent retriever. To streamline the process and elevate the demand for arranging, we utilize LLM to determine the appropriate API to call as part of reasoning.

Our filtered ToolBench dataset consists of 13k training instances and 1.5k test instances. Similar to math plan generation, we prompt Qwen2-7b-Instruct to generate tool calling plans, with tasks $q$ and answer $a$. 

\subsection{Evaluation Metrics \& Baselines}
Consistent with other works in math \cite{wei2022chain,yuan2023scaling}, we assess accuracy of our model based on final answer. Specifically, we utilize the third-party library sympy.parsing.latex\footnote{https://docs.sympy.org/latest/modules/parsing.html} to determine numerical equality.

In assessing the model's performance in tool utilization, we refer to methods used by \citet{chen2024t} and \citet{li2023api}. Our evaluation focuses on comparing the sequences of tool callings generated by the model to corresponding ground-truth sequences. For the evaluation of API selection, we measure f1-score by comparing model's output sequence to the ground-truth sequence. In the case of parameter passing for APIs, we calculate the similarity between parameter lists of the two sequences, under the condition that API selection is accurate. 

Assume the predicted API call list $A=\{API_{ai}\}_{i=1}^m$ and the ground-truth call list $B = \{API_{bj}\}_{j=1}^n$. Suppose the corresponding parameter of $API_{i}$ is $\theta_{i}$, the accuracy obtained by comparing the two sequences can be represented as:
\begin{equation}
	\label{eq5}
	Acc=\frac{\sum\limits_{i=1}^m\sum\limits_{j=1}^nf(API_{ai},API_{bj})sim(\theta_{ai},\theta_{bj})}{m}
\end{equation}
The matching function $f(a,b)$ returns 1 only when $a=b$; otherwise, its value is 0. $Sim(a,b)$ represents the similarity between string $a$ and $b$, which is measured by ROUGE-L. Similar to accuracy calculation in Equation~\ref{eq5}, we compute recall and report the F1-score as final metric of the tool utilization.

We select three models for our experiment: Llama2-7b-chat\cite{touvron2023llama}, Llama3-8b-Instruct\cite{dubey2024llama}, and Qwen2-7b-Instruct\cite{yang2024qwen2}. This selection includes key open-source language models currently available, spanning various capability levels to ensure a degree of representativeness.

We evaluate the performance which used current arranging-executing CoT steps for SFT ($L_{ans}$, Equation~\ref{eq4}). This will serve as the baseline for comparing with our plan-based method, which specifically enhance model's arranging ability.

\subsection{Implementation Details}
We fine-tune our models using LoRA method\cite{hu2021lora}, with the dimensionality of the LoRA low-rank matrix set to 8 and the scaling coefficient lora-alpha of the low-rank matrix set to 32. We train two epochs on all of the models. The learning rate we set for Llama2 is 2e-5, and 2e-6 for Llama3 and Qwen2. We apply AdamW\cite{loshchilov2017decoupled} as optimizer and set warm-up steps as 400. Our reasoning experiments are conducted on TITAN RTX GPUs, and the fine-tuning experiments are conducted on A800 GPUs.

\subsection{Main Results}
\begin{table}
	\centering
	\begin{tabular}{cccc}
		\hline
		\textbf{Model} & \textbf{Original} & \textbf{\begin{tabular}[c]{@{}c@{}}SFT\\ (CoT)\end{tabular}} & \textbf{\begin{tabular}[c]{@{}c@{}}Ours\\ (plan-based)\end{tabular}} \\ \hline
		Llama2         & 27.73           & 28.66                                                        & 31.31                                                      \\
		Llama3         & 78.01           & 72.33                                                        & 77.02                                                      \\
		Qwen2          & 83.93           & 81.05                                                        & 82.11                                                      \\ \hline
	\end{tabular}
	\caption{Accuracy(\%) of models on GSM8k. \textbf{Original} refers to the model without SFT. The baseline \textbf{SFT(CoT)} denotes model trained on current CoT steps. \textbf{Ours(plan-based)} signifies our method focus on reasoning bottleneck enhancement.}
	\label{tab1}
\end{table}

\begin{table}
	\centering
	\begin{tabular}{cccc}
		\hline
		\textbf{Model} & \textbf{Original} & \textbf{\begin{tabular}[c]{@{}c@{}}SFT\\ (CoT)\end{tabular}} & \textbf{\begin{tabular}[c]{@{}c@{}}Ours\\ (plan-based)\end{tabular}} \\ \hline
		Llama2         & 14.86               & 64.73                                                            & 68.04                                                       \\
		Llama3         & 38.30           & 68.56                                                        & 71.88                                                      \\
		Qwen2          & 47.87           & 68.49                                                        & 74.72                                                     \\ \hline
	\end{tabular}
	\caption{The f1-score(\%) of models on ToolBench.}
	\label{tab2}
\end{table}

Table~\ref{tab1} displays the outcomes of fine-tuning and reasoning on math utilizing our method. The results indicate that, compared to directly training on CoT steps containing arranging and executing, our plan-based approach designed to address the arranging bottleneck is more effective in improving performance on math. 

Table~\ref{tab2} presents the outcomes on tool utilization. Mirroring its efficacy in math, our approach has likewise demonstrated substantial improvements in tool utilization. After the fine-tuning of all three models according to our method, the results nearly doubled compared to \textbf{Original}, and surpass the performance of \textbf{SFT(CoT)}. 

The results imply that our plan-based approach is more effective in enhancing the multi-step reasoning ability of models, and alleviating arranging bottleneck to a certain extent in both math and tool tasks.

We also notice that in math task, the performance of Llama3 and Qwen2 after training is not as good as their performance before; while the improvement after fine-tuning on Llama2 is obvious. We speculate that this might be due to the fact that Llama3 and Qwen2 already exhibit commendable accuracy (nearly 80\%), as publishers have utilized relevant data during their training, rendering our secondary fine-tuning efforts less pronounced.

\subsection{Analysis}
We aim to discuss following two research questions (RQs) in this section: (1) How do the components of plan-centric SFT and plan augment reasoning each contribute to our approach? (2) How does our approach specifically alleviate arranging bottleneck of models?
\begin{figure*}[htb]
	\includegraphics[width=1\linewidth]{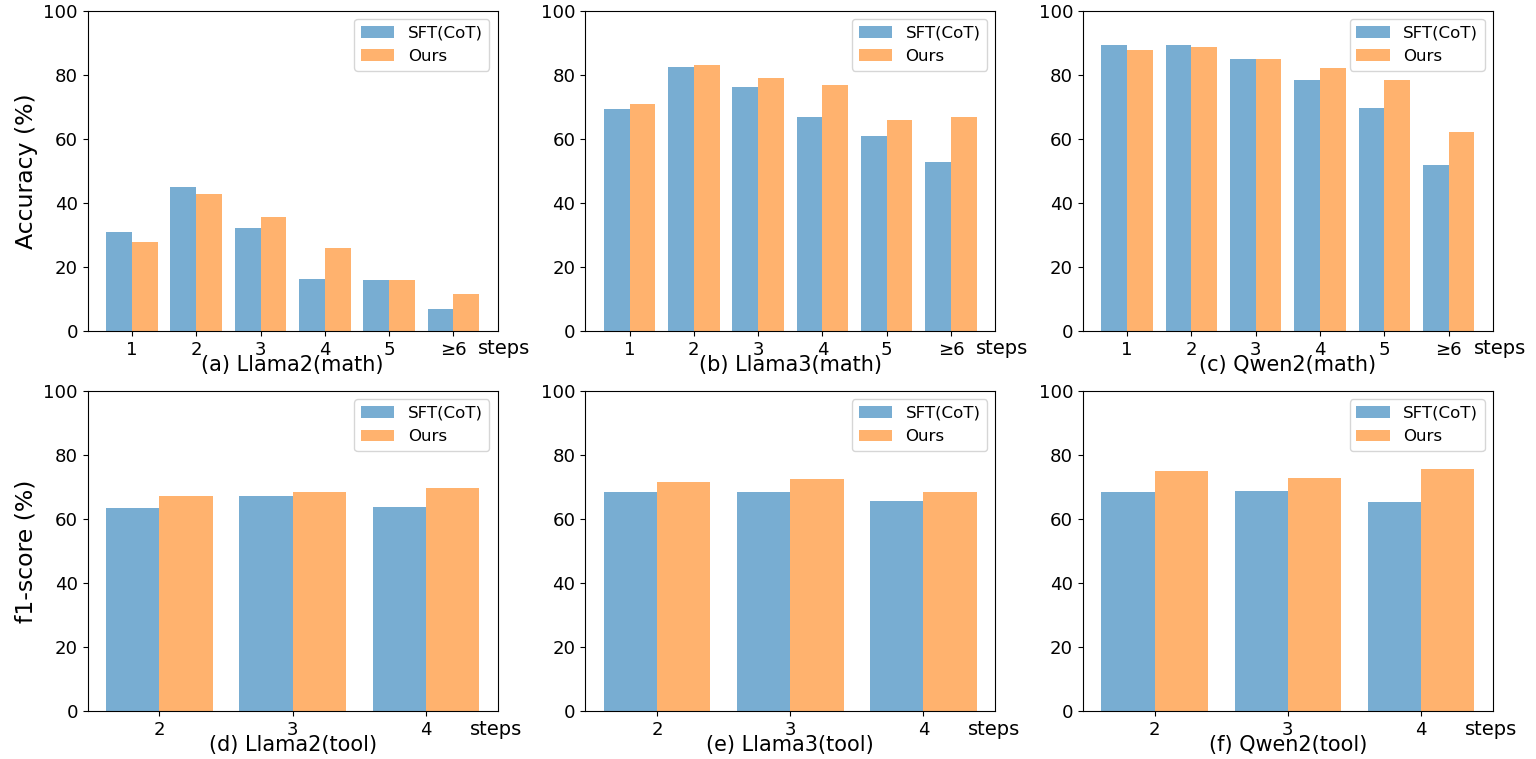} \hfill
	\caption {Score distribution on reasoning steps. Results of math problem are in (a)$\sim$(c), while (d)$\sim$(f) are tool utilization. \textbf{SFT(CoT)} means model trained on CoT steps, and \textbf{Ours} represents our plan-based method. We combine math reasoning steps $\geq6$ together to avoid the impact of small sample size. The improvement on both tasks are significant($p=0.005$ in math, $p=0.001$ in tool utilization).}
	\label{distribution}
\end{figure*}

\noindent \textbf{RQ1: What's the role of each part in our method?} We aim to investigate the role of each part in our method through ablation study here, which include two aspects: (1) How does our plan-centric SFT objective function $L_{all}$ contribute to training stage? (2) How does our plan augment reasoning contribute to reasoning stage? 

\begin{table}[th]
	\centering
	\begin{tabular}{cccc}
		\hline
		\textbf{Model} & \textbf{\begin{tabular}[c]{@{}c@{}}Complete\\Method\end{tabular}} & \textbf{\begin{tabular}[c]{@{}c@{}}w/o training\\ on $L_{ans}$\end{tabular}} & \textbf{\begin{tabular}[c]{@{}c@{}}w/o plan\\ augment\\ reasoning\end{tabular}} \\ \hline
		Llama2         & 31.31                                                      & 31.01                                                                      & 30.78                                                                          \\
		Llama3         & 77.02                                                               & 76.88                                                             & 
		73.46                                                                            \\
		Qwen2          & 82.11                                                      & 83.24                                                                      & 
		80.06                                                                           \\ \hline
	\end{tabular}
	\caption{Comparison among different settings in ablation study, we report the accuracy on GSM8k.}
	\label{ablation-math}
\end{table}

\begin{table}[th]
	\centering
	\begin{tabular}{cccc}
		\hline
			\textbf{Model} & \textbf{\begin{tabular}[c]{@{}c@{}}Complete\\ Method\end{tabular}} & \textbf{\begin{tabular}[c]{@{}c@{}}w/o training\\ on $L_{ans}$\end{tabular}} & \textbf{\begin{tabular}[c]{@{}c@{}}w/o plan\\ augment\\ reasoning\end{tabular}} \\ \hline
		Llama2         & 68.04                                                      & 67.40                                                              & 67.30                                                                         \\
		Llama3         & 71.88                                                               & 62.85                                                             & 69.63                                                                              \\
		Qwen2          & 74.72                                                      & 73.40                                                                      & 73.50                                                                              \\ \hline
	\end{tabular}
	\caption{Comparison among different settings in ablation study, we report the f1-score on ToolBench.}
	\label{ablation-tool}
\end{table}

For the first question, we replace our loss $L_{all}$ with $L_{plan}$ in training objective, while keeping other components unchanged \textbf{(w/o training on $L_{ans}$)}. For the second question, we directly ask the fine-tuned model to reason under CoT, without plan guided \textbf{(w/o plan augment reasoning)}. The results are presented in Table~\ref{ablation-math} and Table~\ref{ablation-tool}.

For question 1, the results demonstrate that for almost all settings, employing the training objective $L_{all}$ leads to better outcomes compared to training with a single loss $L_{plan}$, training on answers can help to improve model's planning ability. For question 2, Our plan-augment reasoning also achieves a better result compared to current CoT method. Results show that by separating abstract plans from arranging-executing steps, the generated plan can guide model to reason along the correct path. We also find that for models like Qwen2 that have been fully trained on math, continuing to train on CoT steps may have negative effects; while training on our abstract plan can alleviate it.

\noindent \textbf{RQ2: How does our approach specifically alleviate arranging bottleneck of the model?} In this context, we aim to further investigate how our plan-based training and reasoning method alleviate arranging bottleneck and the underlying aspects it encompasses. To achieve this, we conduct statistics based on the number of reasoning steps for each question, and the results are presented in Figure~\ref{distribution}.

We find that the performance of current CoT reasoning tends to decline as the number of reasoning steps increases, whereas our method significantly mitigates this degradation trend. This implies that our method is more beneficial on long-distance multi-step reasoning problems, while CoT focus on simple problems with only 2$\sim$3 steps. This conclusion is much more pronounced in the field of math. In tool utilization, the improvement seems more balanced. 

We suggest this might be due to the fact that as the number of reasoning steps increases, the model is more likely to deviate from the correct reasoning path, because the previous generated content will influence the following arranging in self-regressive manner. Our approach, by generating plans to constrain model's arranging space in each step, improves the generalization of models in long-distance multi-step reasoning problems.

\section{Conclusions}
In this research, we conduct an analysis of two crucial aspects of CoT in multi-step reasoning tasks of language models: arranging and executing. We employ both theoretical derivation and empirical study, using math problem as a case study. The results indicate that arranging rather than executing is the primary bottleneck for current models in multi-step reasoning tasks. To address this, we propose a plan-based training and reasoning framework which contains plan augment reasoning and plan-centric SFT, specifically aimed at enhancing arranging ability of language models. Compared to previous methods which directly fine-tuning with CoT steps, our approach yields superior outcomes in both math and tool utilization tasks, particularly exhibits generalization in scenarios involving long-distance reasoning.
\section{Limitations}
Our work still has some limitations. First, we have yet to experiment on other multi-step reasoning tasks beyond math and tool utilization, such as code generation, to further validate the generalization of our method. Second, we still rely on collecting more high-quality data to further improve task's performance. Finally, we look forward to adopting a more diverse set of evaluation metrics and benchmarks to evaluate from different angles. We will continue to improve upon these limitations in our future work.

\bibliography{custom}

\begin{thebibliography}{34}
\providecommand{\natexlab}[1]{#1}

\bibitem[{Achiam et~al.(2023)Achiam, Adler, Agarwal, Ahmad, Akkaya, Aleman,
  Almeida, Altenschmidt, Altman, Anadkat et~al.}]{Achiam2023GPT4TR}
Josh Achiam, Steven Adler, Sandhini Agarwal, Lama Ahmad, Ilge Akkaya,
  Florencia~Leoni Aleman, Diogo Almeida, Janko Altenschmidt, Sam Altman,
  Shyamal Anadkat, et~al. 2023.
\newblock Gpt-4 technical report.
\newblock \emph{arXiv preprint arXiv:2303.08774}.

\bibitem[{Bai et~al.(2023)Bai, Bai, Chu, Cui, Dang, Deng, Fan, Ge, Han, Huang
  et~al.}]{qwen1.5}
Jinze Bai, Shuai Bai, Yunfei Chu, Zeyu Cui, Kai Dang, Xiaodong Deng, Yang Fan,
  Wenbin Ge, Yu~Han, Fei Huang, et~al. 2023.
\newblock Qwen technical report.
\newblock \emph{arXiv preprint arXiv:2309.16609}.

\bibitem[{Besta et~al.(2024)Besta, Blach, Kubicek, Gerstenberger, Podstawski,
  Gianinazzi, Gajda, Lehmann, Niewiadomski, Nyczyk et~al.}]{besta2024graph}
Maciej Besta, Nils Blach, Ales Kubicek, Robert Gerstenberger, Michal
  Podstawski, Lukas Gianinazzi, Joanna Gajda, Tomasz Lehmann, Hubert
  Niewiadomski, Piotr Nyczyk, et~al. 2024.
\newblock Graph of thoughts: Solving elaborate problems with large language
  models.
\newblock In \emph{Proceedings of the AAAI Conference on Artificial
  Intelligence}, volume~38, pages 17682--17690.

\bibitem[{Chen et~al.(2024)Chen, Du, Zhang, Liu, Liu, Zheng, Zhuo, Zhang, Lin,
  Chen et~al.}]{chen2024t}
Zehui Chen, Weihua Du, Wenwei Zhang, Kuikun Liu, Jiangning Liu, Miao Zheng,
  Jingming Zhuo, Songyang Zhang, Dahua Lin, Kai Chen, et~al. 2024.
\newblock T-eval: Evaluating the tool utilization capability of large language
  models step by step.
\newblock In \emph{Proceedings of the 62nd Annual Meeting of the Association
  for Computational Linguistics (Volume 1: Long Papers)}, pages 9510--9529.

\bibitem[{Cobbe et~al.(2021)Cobbe, Kosaraju, Bavarian, Chen, Jun, Kaiser,
  Plappert, Tworek, Hilton, Nakano et~al.}]{cobbe2021training}
Karl Cobbe, Vineet Kosaraju, Mohammad Bavarian, Mark Chen, Heewoo Jun, Lukasz
  Kaiser, Matthias Plappert, Jerry Tworek, Jacob Hilton, Reiichiro Nakano,
  et~al. 2021.
\newblock Training verifiers to solve math word problems.
\newblock \emph{arXiv preprint arXiv:2110.14168}.

\bibitem[{Dubey et~al.(2024)Dubey, Jauhri, Pandey, Kadian, Al-Dahle, Letman,
  Mathur, Schelten, Yang, Fan et~al.}]{dubey2024llama}
Abhimanyu Dubey, Abhinav Jauhri, Abhinav Pandey, Abhishek Kadian, Ahmad
  Al-Dahle, Aiesha Letman, Akhil Mathur, Alan Schelten, Amy Yang, Angela Fan,
  et~al. 2024.
\newblock The llama 3 herd of models.
\newblock \emph{arXiv preprint arXiv:2407.21783}.

\bibitem[{Feng et~al.(2023)Feng, Zhang, Gu, Ye, He, and
  Wang}]{NEURIPS2023_dfc310e8}
Guhao Feng, Bohang Zhang, Yuntian Gu, Haotian Ye, Di~He, and Liwei Wang. 2023.
\newblock \href
  {https://proceedings.neurips.cc/paper_files/paper/2023/file/dfc310e81992d2e4cedc09ac47eff13e-Paper-Conference.pdf}
  {Towards revealing the mystery behind chain of thought: A theoretical
  perspective}.
\newblock In \emph{Advances in Neural Information Processing Systems},
  volume~36, pages 70757--70798. Curran Associates, Inc.

\bibitem[{Guo et~al.(2024)Guo, Cheng, Wang, Liang, Qin, Li, Liu, Sun, and
  Liu}]{guo2024stabletoolbench}
Zhicheng Guo, Sijie Cheng, Hao Wang, Shihao Liang, Yujia Qin, Peng Li, Zhiyuan
  Liu, Maosong Sun, and Yang Liu. 2024.
\newblock Stabletoolbench: Towards stable large-scale benchmarking on tool
  learning of large language models.
\newblock \emph{arXiv preprint arXiv:2403.07714}.

\bibitem[{Hu et~al.(2021)Hu, Shen, Wallis, Allen-Zhu, Li, Wang, Wang, and
  Chen}]{hu2021lora}
Edward~J Hu, Yelong Shen, Phillip Wallis, Zeyuan Allen-Zhu, Yuanzhi Li, Shean
  Wang, Lu~Wang, and Weizhu Chen. 2021.
\newblock Lora: Low-rank adaptation of large language models.
\newblock \emph{arXiv preprint arXiv:2106.09685}.

\bibitem[{Javaheripi et~al.(2023)Javaheripi, Bubeck, Abdin, Aneja, Bubeck,
  Mendes, Chen, Del~Giorno, Eldan, Gopi et~al.}]{javaheripi2023phi}
Mojan Javaheripi, S{\'e}bastien Bubeck, Marah Abdin, Jyoti Aneja, Sebastien
  Bubeck, Caio C{\'e}sar~Teodoro Mendes, Weizhu Chen, Allie Del~Giorno, Ronen
  Eldan, Sivakanth Gopi, et~al. 2023.
\newblock Phi-2: The surprising power of small language models.
\newblock \emph{Microsoft Research Blog}.

\bibitem[{Kaplan et~al.(2020)Kaplan, McCandlish, Henighan, Brown, Chess, Child,
  Gray, Radford, Wu, and Amodei}]{Kaplan2020ScalingLF}
Jared Kaplan, Sam McCandlish, Tom Henighan, Tom~B. Brown, Benjamin Chess, Rewon
  Child, Scott Gray, Alec Radford, Jeff Wu, and Dario Amodei. 2020.
\newblock \href {https://api.semanticscholar.org/CorpusID:210861095} {Scaling
  laws for neural language models}.
\newblock \emph{ArXiv}, abs/2001.08361.

\bibitem[{Li et~al.(2024)Li, Wang, Hu, Wei, Zheng, Hu, Zhang, and
  Peng}]{li2024common}
Chen Li, Weiqi Wang, Jingcheng Hu, Yixuan Wei, Nanning Zheng, Han Hu, Zheng
  Zhang, and Houwen Peng. 2024.
\newblock Common 7b language models already possess strong math capabilities.
\newblock \emph{arXiv preprint arXiv:2403.04706}.

\bibitem[{Li et~al.(2023{\natexlab{a}})Li, Li, Li, and
  Jin}]{li2023structuredchainofthoughtpromptingcode}
Jia Li, Ge~Li, Yongmin Li, and Zhi Jin. 2023{\natexlab{a}}.
\newblock \href {https://arxiv.org/abs/2305.06599} {Structured chain-of-thought
  prompting for code generation}.
\newblock \emph{Preprint}, arXiv:2305.06599.

\bibitem[{Li et~al.(2023{\natexlab{b}})Li, Zhao, Yu, Song, Li, Yu, Li, Huang,
  and Li}]{li2023api}
Minghao Li, Yingxiu Zhao, Bowen Yu, Feifan Song, Hangyu Li, Haiyang Yu, Zhoujun
  Li, Fei Huang, and Yongbin Li. 2023{\natexlab{b}}.
\newblock Api-bank: A comprehensive benchmark for tool-augmented llms.
\newblock \emph{arXiv preprint arXiv:2304.08244}.

\bibitem[{Liang et~al.(2023)Liang, Yu, Rajpurohit, Clark, Zhang, and
  Kalyan}]{liang-etal-2023-gpt}
Zhenwen Liang, Wenhao Yu, Tanmay Rajpurohit, Peter Clark, Xiangliang Zhang, and
  Ashwin Kalyan. 2023.
\newblock \href {https://doi.org/10.18653/v1/2023.emnlp-main.889} {Let {GPT} be
  a math tutor: Teaching math word problem solvers with customized exercise
  generation}.
\newblock In \emph{Proceedings of the 2023 Conference on Empirical Methods in
  Natural Language Processing}, pages 14384--14396, Singapore. Association for
  Computational Linguistics.

\bibitem[{Loshchilov and Hutter(2017)}]{loshchilov2017decoupled}
Ilya Loshchilov and Frank Hutter. 2017.
\newblock Decoupled weight decay regularization.
\newblock \emph{arXiv preprint arXiv:1711.05101}.

\bibitem[{Luo et~al.(2023)Luo, Sun, Xu, Zhao, Lou, Tao, Geng, Lin, Chen, and
  Zhang}]{luo2023wizardmath}
Haipeng Luo, Qingfeng Sun, Can Xu, Pu~Zhao, Jianguang Lou, Chongyang Tao, Xiubo
  Geng, Qingwei Lin, Shifeng Chen, and Dongmei Zhang. 2023.
\newblock Wizardmath: Empowering mathematical reasoning for large language
  models via reinforced evol-instruct.
\newblock \emph{arXiv preprint arXiv:2308.09583}.

\bibitem[{Magister et~al.(2023)Magister, Mallinson, Adamek, Malmi, and
  Severyn}]{magister-etal-2023-teaching}
Lucie~Charlotte Magister, Jonathan Mallinson, Jakub Adamek, Eric Malmi, and
  Aliaksei Severyn. 2023.
\newblock \href {https://doi.org/10.18653/v1/2023.acl-short.151} {Teaching
  small language models to reason}.
\newblock In \emph{Proceedings of the 61st Annual Meeting of the Association
  for Computational Linguistics (Volume 2: Short Papers)}, pages 1773--1781,
  Toronto, Canada. Association for Computational Linguistics.

\bibitem[{Momennejad et~al.(2024)Momennejad, Hasanbeig, Vieira~Frujeri, Sharma,
  Jojic, Palangi, Ness, and Larson}]{momennejad2024evaluating}
Ida Momennejad, Hosein Hasanbeig, Felipe Vieira~Frujeri, Hiteshi Sharma,
  Nebojsa Jojic, Hamid Palangi, Robert Ness, and Jonathan Larson. 2024.
\newblock Evaluating cognitive maps and planning in large language models with
  cogeval.
\newblock \emph{Advances in Neural Information Processing Systems}, 36.

\bibitem[{Qiao et~al.(2024)Qiao, Zhang, Fang, Luo, Zhou, Jiang, Lv, and
  Chen}]{qiao2024autoact}
Shuofei Qiao, Ningyu Zhang, Runnan Fang, Yujie Luo, Wangchunshu Zhou,
  Yuchen~Eleanor Jiang, Chengfei Lv, and Huajun Chen. 2024.
\newblock Autoact: Automatic agent learning from scratch via self-planning.
\newblock \emph{arXiv preprint arXiv:2401.05268}.

\bibitem[{Schick et~al.(2024)Schick, Dwivedi-Yu, Dess{\`\i}, Raileanu, Lomeli,
  Hambro, Zettlemoyer, Cancedda, and Scialom}]{schick2024toolformer}
Timo Schick, Jane Dwivedi-Yu, Roberto Dess{\`\i}, Roberta Raileanu, Maria
  Lomeli, Eric Hambro, Luke Zettlemoyer, Nicola Cancedda, and Thomas Scialom.
  2024.
\newblock Toolformer: Language models can teach themselves to use tools.
\newblock \emph{Advances in Neural Information Processing Systems}, 36.

\bibitem[{Sun et~al.(2024)Sun, Chai, Yang, Yin, Guo, Liu, Wang, Yang, and
  Li}]{sun2024unicoder}
Tao Sun, Linzheng Chai, Jian Yang, Yuwei Yin, Hongcheng Guo, Jiaheng Liu, Bing
  Wang, Liqun Yang, and Zhoujun Li. 2024.
\newblock Unicoder: Scaling code large language model via universal code.
\newblock \emph{arXiv preprint arXiv:2406.16441}.

\bibitem[{Touvron et~al.(2023)Touvron, Martin, Stone, Albert, Almahairi,
  Babaei, Bashlykov, Batra, Bhargava, Bhosale et~al.}]{touvron2023llama}
Hugo Touvron, Louis Martin, Kevin Stone, Peter Albert, Amjad Almahairi, Yasmine
  Babaei, Nikolay Bashlykov, Soumya Batra, Prajjwal Bhargava, Shruti Bhosale,
  et~al. 2023.
\newblock Llama 2: Open foundation and fine-tuned chat models.
\newblock \emph{arXiv preprint arXiv:2307.09288}.

\bibitem[{Valmeekam et~al.(2023)Valmeekam, Marquez, Sreedharan, and
  Kambhampati}]{valmeekam2023planning}
Karthik Valmeekam, Matthew Marquez, Sarath Sreedharan, and Subbarao
  Kambhampati. 2023.
\newblock On the planning abilities of large language models-a critical
  investigation.
\newblock \emph{Advances in Neural Information Processing Systems},
  36:75993--76005.

\bibitem[{Wang et~al.(2022{\natexlab{a}})Wang, Wei, Schuurmans, Le, Chi,
  Narang, Chowdhery, and Zhou}]{wang2022self}
Xuezhi Wang, Jason Wei, Dale Schuurmans, Quoc Le, Ed~Chi, Sharan Narang,
  Aakanksha Chowdhery, and Denny Zhou. 2022{\natexlab{a}}.
\newblock Self-consistency improves chain of thought reasoning in language
  models.
\newblock \emph{arXiv preprint arXiv:2203.11171}.

\bibitem[{Wang et~al.(2022{\natexlab{b}})Wang, Wei, Schuurmans, Le, hsin Chi,
  and Zhou}]{Wang2022SelfConsistencyIC}
Xuezhi Wang, Jason Wei, Dale Schuurmans, Quoc Le, Ed~Huai hsin Chi, and Denny
  Zhou. 2022{\natexlab{b}}.
\newblock \href {https://api.semanticscholar.org/CorpusID:247595263}
  {Self-consistency improves chain of thought reasoning in language models}.
\newblock \emph{ArXiv}, abs/2203.11171.

\bibitem[{Wei et~al.(2022)Wei, Wang, Schuurmans, Bosma, Xia, Chi, Le, Zhou
  et~al.}]{wei2022chain}
Jason Wei, Xuezhi Wang, Dale Schuurmans, Maarten Bosma, Fei Xia, Ed~Chi, Quoc~V
  Le, Denny Zhou, et~al. 2022.
\newblock Chain-of-thought prompting elicits reasoning in large language
  models.
\newblock \emph{Advances in neural information processing systems},
  35:24824--24837.

\bibitem[{Xu et~al.(2024)Xu, Hua, Li, and Wang}]{xu-etal-2024-exploring}
Yanzhi Xu, Yueying Hua, Shichen Li, and Zhongqing Wang. 2024.
\newblock \href {https://aclanthology.org/2024.acl-long.6} {Exploring
  chain-of-thought for multi-modal metaphor detection}.
\newblock In \emph{Proceedings of the 62nd Annual Meeting of the Association
  for Computational Linguistics (Volume 1: Long Papers)}, pages 91--101,
  Bangkok, Thailand. Association for Computational Linguistics.

\bibitem[{Yang et~al.(2024)Yang, Yang, Hui, Zheng, Yu, Zhou, Li, Li, Liu, Huang
  et~al.}]{yang2024qwen2}
An~Yang, Baosong Yang, Binyuan Hui, Bo~Zheng, Bowen Yu, Chang Zhou, Chengpeng
  Li, Chengyuan Li, Dayiheng Liu, Fei Huang, et~al. 2024.
\newblock Qwen2 technical report.
\newblock \emph{arXiv preprint arXiv:2407.10671}.

\bibitem[{Yang et~al.(2023)Yang, Ding, Lv, Jiang, He, Guo, Bai, and
  Tang}]{yang2023gpt}
Zhen Yang, Ming Ding, Qingsong Lv, Zhihuan Jiang, Zehai He, Yuyi Guo, Jinfeng
  Bai, and Jie Tang. 2023.
\newblock Gpt can solve mathematical problems without a calculator.
\newblock \emph{arXiv preprint arXiv:2309.03241}.

\bibitem[{Yao et~al.(2024)Yao, Yu, Zhao, Shafran, Griffiths, Cao, and
  Narasimhan}]{yao2024}
Shunyu Yao, Dian Yu, Jeffrey Zhao, Izhak Shafran, Tom Griffiths, Yuan Cao, and
  Karthik Narasimhan. 2024.
\newblock Tree of thoughts: Deliberate problem solving with large language
  models.
\newblock \emph{Advances in Neural Information Processing Systems}, 36.

\bibitem[{Yuan et~al.(2023)Yuan, Yuan, Li, Dong, Lu, Tan, Zhou, and
  Zhou}]{yuan2023scaling}
Zheng Yuan, Hongyi Yuan, Chengpeng Li, Guanting Dong, Keming Lu, Chuanqi Tan,
  Chang Zhou, and Jingren Zhou. 2023.
\newblock Scaling relationship on learning mathematical reasoning with large
  language models.
\newblock \emph{arXiv preprint arXiv:2308.01825}.

\bibitem[{Zhang et~al.(2024)Zhang, Guo, Le, Yang, Liu, Li, Zheng, Xu, Zang,
  Zheng, and Zhang}]{zhang2024lemurlogparsingentropy}
Wei Zhang, Hongcheng Guo, Anjie Le, Jian Yang, Jiaheng Liu, Zhoujun Li, Tieqiao
  Zheng, Shi Xu, Runqiang Zang, Liangfan Zheng, and Bo~Zhang. 2024.
\newblock \href {https://arxiv.org/abs/2402.18205} {Lemur: Log parsing with
  entropy sampling and chain-of-thought merging}.
\newblock \emph{Preprint}, arXiv:2402.18205.

\bibitem[{Zhang et~al.(2022)Zhang, Zhang, Li, and
  Smola}]{zhang2022automaticchainthoughtprompting}
Zhuosheng Zhang, Aston Zhang, Mu~Li, and Alex Smola. 2022.
\newblock \href {https://arxiv.org/abs/2210.03493} {Automatic chain of thought
  prompting in large language models}.
\newblock \emph{Preprint}, arXiv:2210.03493.

\end{thebibliography}

\appendix

\section{Prompt Settings}
\label{appendix1}
\subsection{Prompts for Math Reasoning}
We present all the prompts used in this paper to solve mathematical problems with LLMs. Our prompts are zero-shot and concise, aiming to accurately reflect models' ability while minimizing the influence of prompt engineering. 

When using language models to solve math problems directly, our prompt is:\\
\textbf{You are a math expert and complete the following math problem, let's do it step by step. \\
<Math problem>: \{question\}\\
Your answer should be end with "The final answer is:(your answer)".}

When using the model to generate solving plans, our prompt is:\\
\textbf{You are a math teacher, please give the problem-solving ideas for the following math problem, but you cannot directly give specific mathematical expressions, answers, etc.\\ Your answer format is "To solve this question, first we should...then..."\\
<Math problem>: \{question\}\\
<Idea>:}

When using generated plans to guide models to solve the math problems, our prompt is:\\
\textbf{Solve the <math problem> with the help of the given <idea> by a teacher, which may help you to solve it correctly:\\
<Math problem>: \{question\}\\
<Idea>: \{idea\}\\
Your answer should be end with "The final answer is:(your answer)".Let's think step by step.}

\subsection{Prompts for Tool Utilization}
We present the prompts for our experiment in tool utilization here.

When using language models to complete the tool utilization task directly, our prompt is:\\
\textbf{You are an AutoGPT, capable of utilizing numerous tool-functions to complete the given task. First, I will provide you the task description <Task> with tool-functions <Func\_list> you can use. You have to select several functions from <Func\_list> to finish the task.\\ Your answer format is: \\
step1: [func]function\_name1(parm1=value1, parm2=value2...)[/func]\\
step2: [func]function\_name2(parm1=value1)[/func]\\
...stepN:...\\
<Task>: \{task\}\\
<Func\_list>: \{func\}}

When using language models to generate the solving plan of tool utilization task directly, our prompt is:\\
\textbf{You are an AutoGPT, capable of utilizing numerous tool-functions to complete the given task. I will provide you the task description <Task> with tool-functions <Func\_list> you can use. Please generate the abstract solving idea of the task, which includes the APIs you use from <Func\_list>.\\
<Task>: \{task\}\\
<Func\_list>: \{func\}}

When using the generated plans to guide models to complete tool utilization task, our prompt is:\\
\textbf{You are an AutoGPT, capable of utilizing numerous tool-functions to complete the given task. First, I will provide you the task description <Task> with tool-functions <Func\_list> you can use. You have to select several functions from <Func\_list> to finish the task. You can refer <Idea> to the ideas in which he can help you solve the problem. \\
Your answer format is: \\
step1: [func]function\_name1(parm1=value1, parm2=value2...)[/func]\\
step2: [func]function\_name2(parm1=value1)[/func]\\
...stepN:...\\
<Task>: \{task\}\\
<Func\_list>: \{func\}\\
<Idea>: \{idea\}}
\end{document}